\documentclass[11pt,a4paper]{article}
\usepackage[hyperref]{acl2018}
\usepackage{times}
\usepackage{latexsym}
\usepackage{amsmath}
\usepackage{amssymb}
\usepackage{multirow,makecell}
\usepackage{graphicx}
\usepackage[utf8]{inputenc}
\usepackage{booktabs}
\usepackage{pifont}
\newcommand{\cmark}{\ding{51}}%
\newcommand{\xmark}{\ding{55}}%
\usepackage{breqn}

\usepackage{url}

\aclfinalcopy 
\def\aclpaperid{1305} 

\title{Constituency Parsing with a Self-Attentive Encoder}

\author{Nikita Kitaev \and Dan Klein \\
  Computer Science Division \\
  University of California, Berkeley \\
  {\tt \{kitaev, klein\}@cs.berkeley.edu}}

\date{}

\begin{document}
\maketitle
\begin{abstract}
We demonstrate that replacing an LSTM encoder with a self-attentive architecture can lead to improvements to a state-of-the-art discriminative constituency parser. The use of attention makes explicit the manner in which information is propagated between different locations in the sentence, which we use to both analyze our model and propose potential improvements.  For example, we find that separating positional and content information in the encoder can lead to improved parsing accuracy. Additionally, we evaluate different approaches for lexical representation. Our parser achieves new state-of-the-art results for single models trained on the Penn Treebank: 93.55 F1 without the use of any external data, and 95.13 F1 when using pre-trained word representations. Our parser also outperforms the previous best-published accuracy figures on 8 of the 9 languages in the SPMRL dataset.
\end{abstract}

\section{Introduction}
\label{sec:intro}

\begin{figure}
  \centering
    \includegraphics[width=0.35\textwidth]{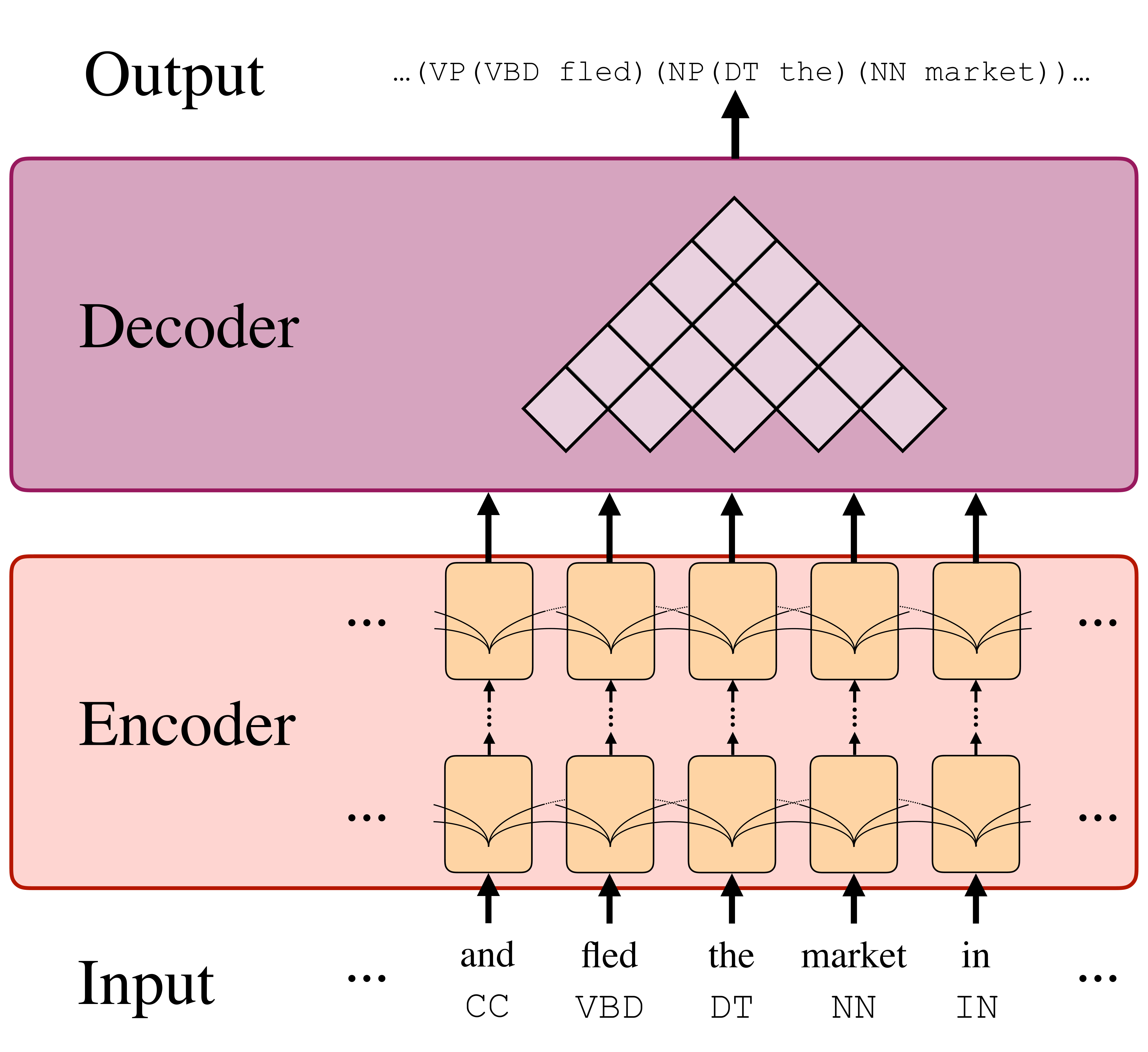}
    \caption{Our parser combines a chart decoder with a sentence encoder based on self-attention.}
    \label{fig:encdec}
\end{figure}

In recent years, neural network approaches have led to improvements in constituency parsing \citep{dyer_recurrent_2016,cross_span-based_2016,choe_parsing_2016,stern_minimal_2017,fried_improving_2017}. Many of these parsers can broadly be characterized as following an encoder-decoder design: an encoder reads the input sentence and summarizes it into a vector or set of vectors (e.g.\ one for each word or span in the sentence), and then a decoder uses these vector summaries to incrementally build up a labeled parse tree. In contrast to the large variety of decoder architectures investigated in recent work, the encoders in recent parsers have predominantly been built using recurrent neural networks (RNNs), and in particular Long Short-Term Memory networks (LSTMs). RNNs have largely replaced approaches such as the fixed-window-size feed-forward networks of \citet{durrett_neural_2015} in part due to their ability to capture global context. However, RNNs are not the only architecture capable of summarizing large global contexts: recent work by \citet{vaswani_attention_2017} presented a new state-of-the-art approach to machine translation with an architecture that entirely eliminates recurrent connections and relies instead on a repeated neural attention mechanism. In this paper, we introduce a parser that combines an encoder built using this kind of self-attentive architecture with a decoder customized for parsing (Figure~\ref{fig:encdec}). In Section~\ref{sec:base-model} of this paper, we describe the architecture and present our finding that self-attention can outperform an LSTM-based approach.

A neural attention mechanism makes explicit the manner in which information is transferred between different locations in the sentence, which we can use to study the relative importance of different kinds of context to the parsing task. Different locations in the sentence can attend to each other based on their positions, but also based on their contents (i.e.\ based on the words at or around those positions). In Section~\ref{sec:factored-model} we present our finding that when our parser learns to make an implicit trade-off between these two types of attention, it predominantly makes use of position-based attention, and show that explicitly factoring the two types of attention can noticeably improve parsing accuracy. In Section~\ref{sec:analysis}, we study our model's use of attention and reaffirm the conventional wisdom that sentence-wide global context is important for parsing decisions.

Like in most neural parsers, we find morphological (or at least sub-word) features to be important to achieving good results, particularly on unseen words or inflections. In Section~\ref{subsec:morpho}, we demonstrate that a simple scheme based on concatenating character embeddings of word prefixes/suffixes can outperform using part-of-speech tags from an external system. We also present a version of our model that uses a character LSTM, which performs better than other lexical representations -- even if word embeddings are removed from the model. In Section~\ref{subsec:elmo}, we explore an alternative approach for lexical representations that makes use of pre-training on a large unsupervised corpus. We find that using the deep contextualized representations proposed by \citet{peters_deep_2018} can boost parsing accuracy.

Our parser achieves 93.55 F1 on the Penn Treebank WSJ test set when not using external word representations, outperforming all previous single-system constituency parsers trained only on the WSJ training set. The addition of pre-trained word representations following \citet{peters_deep_2018} increases parsing accuracy to 95.13 F1, a new state-of-the-art for this dataset. Our model also outperforms previous best published results on 8 of the 9 languages in the SPMRL 2013/2014 shared tasks. Code and trained English models are publicly available.\footnote{\href{https://github.com/nikitakit/self-attentive-parser}{https://github.com/nikitakit/self-attentive-parser}}

\section{Base Model}
\label{sec:base-model}

Our parser follows an encoder-decoder architecture, as shown in Figure~\ref{fig:encdec}. The decoder, described in Section~\ref{subsec:decoder}, is borrowed from the chart parser of \citet{stern_minimal_2017} with additional modifications from \citet{gaddy_analysis_2018}. Their parser is architecturally streamlined yet achieves the highest performance among discriminative single-system parsers trained on WSJ data only, which is why we selected it as the starting point for our experiments with encoder variations. Sections~\ref{subsec:linear-encoder} and \ref{subsec:span-encoder} describe the base version of our encoder, where the self-attentive architecture described in Section~\ref{subsec:linear-encoder} is adapted from \citet{vaswani_attention_2017}.

\subsection{Tree Scores and Chart Decoder}
\label{subsec:decoder}

Our parser assigns a real-valued score $s(T)$ to each tree $T$, which decomposes as \begin{equation} s(T) = \sum_{(i,j,l)\in T} s(i,j,l) \label{eqn:tree-score}\end{equation}
Here $s(i,j,l)$ is a real-valued score for a constituent that is located between fencepost positions $i$ and $j$ in a sentence and has label $l$. To handle unary chains, the set of labels includes a collapsed entry for each unary chain in the training set. The model handles $n$-ary trees by binarizing them and introducing a dummy label $\varnothing$ to nodes created during binarization, with the property that $ \forall i,j: s(i,j,\varnothing)=0$. Enforcing that scores associated with the dummy labels are always zero ensures that (\ref{eqn:tree-score}) continues to hold for all possible binarizations of an $n$-ary tree.

At test time, the model-optimal tree $$\hat{T} = \arg\max_{T} s(T)$$ can be found efficiently using a CKY-style inference algorithm. Given the correct tree $T^\star$, the model is trained to satisfy the margin constraints $$s(T^\star) \geq s(T) + \Delta(T, T^\star)$$ for all trees $T$ by minimizing the hinge loss $$\max \Big(0, \max_{T \neq T^\star} \big[s(T) + \Delta(T, T^\star) \big] - s(T^\star)  \Big)$$
Here $\Delta$ is the Hamming loss on labeled spans, and the tree corresponding to the most-violated constraint can be found using a slight modification of the inference algorithm used at test time.

For further details, see \citet{gaddy_analysis_2018}. The remainder of this paper concerns itself with the functional form of $s(i,j,l)$, which is calculated using a neural network for all $l \neq \varnothing$.

\subsection{Context-Aware Word Representations}
\label{subsec:linear-encoder}

The encoder portion of our model is split into two parts: a word-based portion that assigns a context-aware vector representation $y_t$ to each position $t$ in the sentence (described in this section), and a chart portion that combines the vectors $y_t$ to generate span scores $s(i,j,l)$ (Section~\ref{subsec:span-encoder}). The architecture for generating the vectors $y_t$ is adapted from \citet{vaswani_attention_2017}.

\begin{figure}
  \centering
    \includegraphics[width=0.4\textwidth]{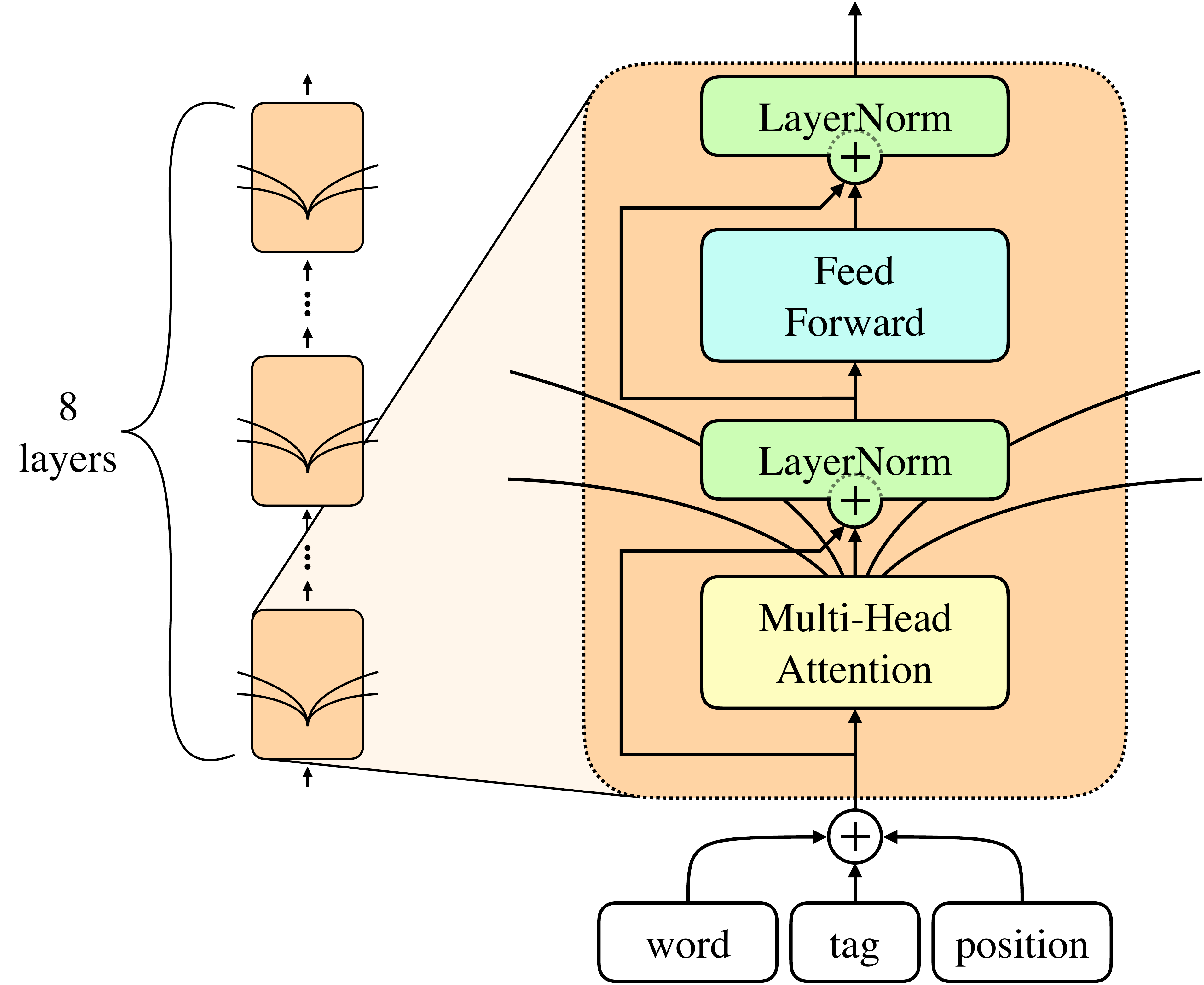}
    \caption{An overview of our encoder, which produces a context-aware summary vector for each word in the sentence. The multi-headed attention mechanism is the only means by which information may propagate between different positions in the sentence.}
    \label{fig:linear-encoder}
\end{figure}

The encoder takes as input a sequence of word embeddings $[w_1, w_2, \ldots, w_T]$, where the first and last embeddings are of special \emph{start} and \emph{stop} tokens. All word embeddings are learned jointly with other parts of the model. To better generalize to words that are not seen during training, the encoder also receives a sequence of part-of-speech tag embeddings $[m_1, m_2, \ldots, m_T]$ based on the output of an external tagger (alternative lexical representations are discussed in Section~\ref{sec:lexical}). Additionally, the encoder stores a learned table of position embeddings, where every number $i \in 1,2,\ldots$ (up to some maximum sentence length) is associated with a vector $p_i$. All embeddings have the same dimensionality, which we call $d_{model}$, and are added together at the input of the encoder: $z_t = w_t + m_t + p_t$.

The vectors $[z_1, z_2, \ldots, z_T]$ are transformed by a stack of 8 identical layers, as shown in Figure~\ref{fig:linear-encoder}. Each layer consists of two stacked sublayers: a multi-headed attention mechanism and a position-wise feed-forward sublayer. The output of each sublayer given an input $x$ is $\mathrm{LayerNorm}(x + \mathrm{SubLayer}(x))$, i.e. each sublayer is followed by a residual connection and a Layer Normalization \citep{ba_layer_2016} step.
As a result, all sublayer outputs, including final outputs $y_t$, are of size $d_{model}$.

\subsubsection{Self-Attention}
\label{subsec:self-attention}

The first sublayer in each of our 8 layers is a multi-headed self-attention mechanism, which is the only means by which information may propagate between positions in the sentence. The input to the attention mechanism is a $T\times d_{model}$ matrix $X$, where each row vector $x_t$ corresponds to word $t$ in the sentence.

\begin{figure}
  \centering
    \includegraphics[width=0.45\textwidth]{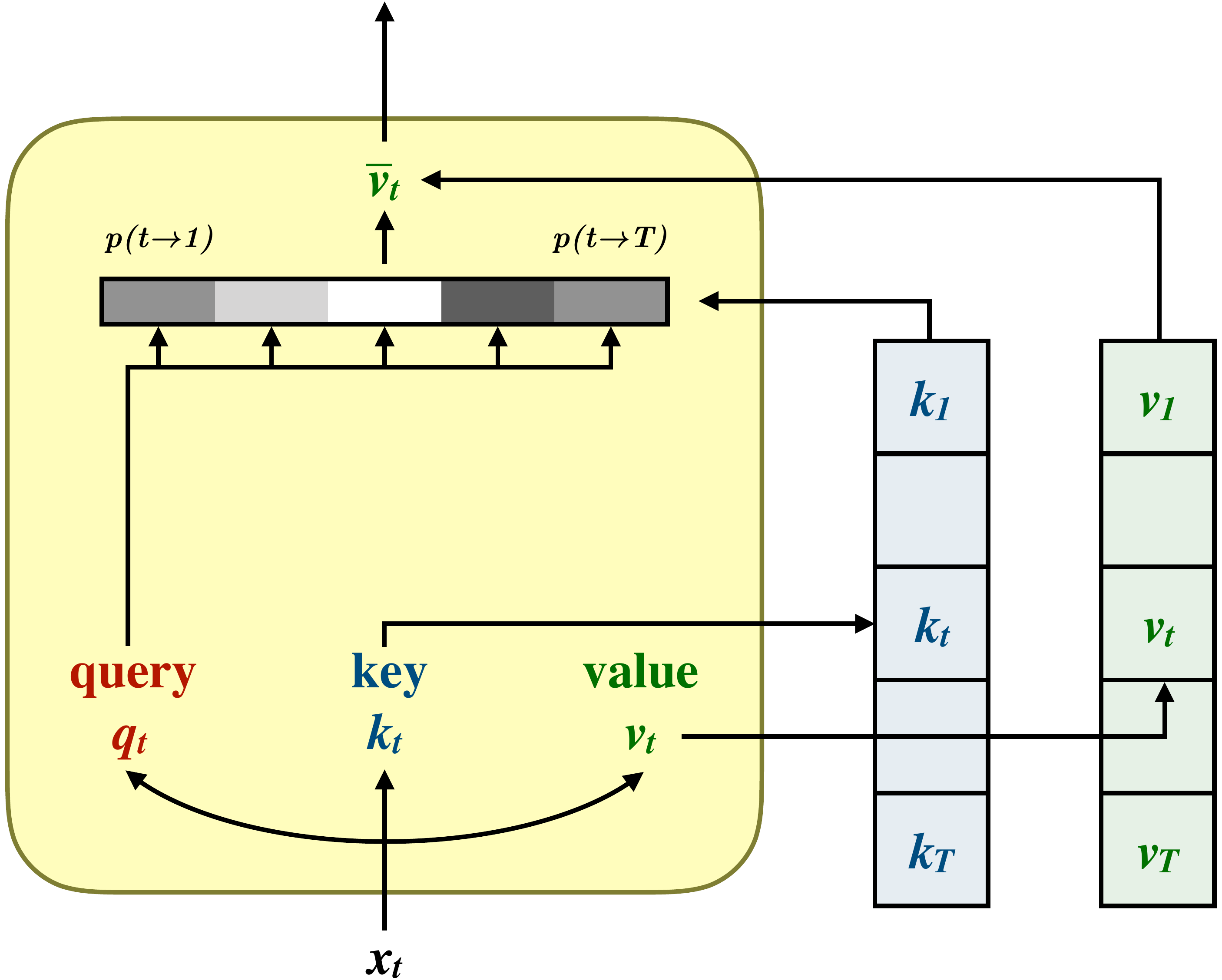}
    \caption{A single attention head. An input $x_t$ is split into three vectors that participate in the attention mechanism: a query $q_t$, a key $k_t$, and a value $v_t$. The query $q_t$ is compared with all keys to form a probability distribution $p(t \rightarrow \cdot)$, which is then used to retrieve an average value $\bar{v}_t$.}
    \label{fig:unfactored-head}
\end{figure}

We first consider a single attention head, as illustrated in Figure~\ref{fig:unfactored-head}. Learned parameter matrices $W_Q$, $W_K$, and $W_V$ are used to map an input $x_t$ to three vectors: a query $q_t=W_Q^\top x_t$, a key $k_t=W_K^\top x_t$, and a value $v_t=W_V^\top x_t$. Query and key vectors have the same number of dimensions, which we call $d_k$. The probability that word $i$ attends to word $j$ is then calculated as $p(i\rightarrow j) \propto \exp(\frac{q_i\cdot k_j}{\sqrt{d_k}})$. The values $v_j$ for all words that have been attended to are aggregated to form an average value $\bar{v}_i = \sum_j p(i\rightarrow j) v_j$, which is projected back to size $d_{model}$ using a learned matrix $W_O$. In matrix form, the behavior of a single attention head is:
\begin{gather*}
\mathrm{SingleHead}(X) = \left[\mathrm{Softmax}\left(\frac{QK^\top}{\sqrt{d_k}}\right)V\right] W_O \\
\text{where}~ Q = XW_Q;~ K = XW_K;~ V = XW_V
\label{eqn:qkvmatrices}
\end{gather*}

Rather than using a single head, our model sums together the outputs from multiple heads:
\begin{gather*}
\mathrm{MultiHead}(X) = \sum_{n=1}^{8} \mathrm{SingleHead}^{(n)}(X)
\end{gather*}
Each of the 8 heads has its own trainable parameters $W_Q^{(n)}$, $W_K^{(n)}$, $W_V^{(n)}$, and $W_O^{(n)}$. This allows a word to gather information from up to 8 remote locations in the sentence at each attention sublayer.

\subsubsection{Position-Wise Feed-Forward Sublayer}
\label{subsec:feed-forward}

We use the same form as \citet{vaswani_attention_2017}: $$\mathrm{FeedForward}(x) = W_2 \text{relu}(W_1x + b_1) + b_2$$
Here \textit{relu} denotes the Rectified Linear Unit nonlinearity, and distinct sets of learned parameters are used at each of the 8 instances of the feed-forward sublayer in our model.

The input and output dimensions are the same because of the use of residual connections throughout the model, but we can vary the number of parameters by adjusting the size of the intermediate vector that the nonlinearity is applied to.

\subsection{Span Scores}
\label{subsec:span-encoder}

The outputs $y_t$ from the word-based encoder portion described in the previous section are combined to form span scores $s(i,j,\cdot)$ following the method of \citet{stern_minimal_2017}.
Concretely, $$s(i,j,\cdot) = M_2 \text{relu}(\mathrm{LayerNorm}(M_1v + c_1)) + c_2$$
where $\mathrm{LayerNorm}$ denotes Layer Normalization, \emph{relu} is the Rectified Linear Unit nonlinearity, and $v = [\overset{\rightarrow}{y}_{j} - \overset{\rightarrow}{y}_{i}; \overset{\leftarrow}{y}_{j+1} - \overset{\leftarrow}{y}_{i+1}]$ combines summary vectors for relevant positions in the sentence. A span endpoint to the right of the word potentially requires different information from the endpoint to the left, so a word at a position $k$ is associated with \emph{two} annotation vectors ($\overset{\rightarrow}{y}_{k}$ and $\overset{\leftarrow}{y}_{k}$). 

\citet{stern_minimal_2017} define $\overset{\rightarrow}{y}_{k}$ and $\overset{\leftarrow}{y}_{k}$ in terms of the output of the forward and backward portions, respectively, of their BiLSTM encoder; we instead construct each of $\overset{\rightarrow}{y}_{k}$ and $\overset{\leftarrow}{y}_{k}$ by splitting in half\footnote{To avoid an adverse interaction with material described in Section~\ref{sec:factored-model}, when a vector $y_k$ is split in half the even coordinates contribute to $\overset{\rightarrow}{y}_{k}$ and the odd coordinates contribute to $\overset{\leftarrow}{y}_{k}$.} the outputs $y_k$ from Section~\ref{subsec:linear-encoder}. We also introduce a Layer Normalization step to match the use of Layer Normalization throughout our model. 

\subsection{Results}

The model presented above achieves a score of 92.67 F1 on the Penn Treebank WSJ development set. Details regarding hyperparameter choice and optimizer settings are presented in Appendix~\ref{sec:hyperparams}.
For comparison, a model that uses the same decode procedure with an LSTM-based encoder achieves a development set score of 92.24 \citep{gaddy_analysis_2018}. These results demonstrate that an RNN-based encoder is not required for building a good parser; in fact, self-attention can achieve better results.
  
\section{Content vs. Position Attention}
\label{sec:factored-model}

The primary mechanism for information transfer throughout our encoder is self-attention, where words can attend to each other using both content features and position information. In Section~\ref{sec:base-model}, we described an encoder that takes as input a component-wise addition between a word, tag, and position embedding for each word in the sentence. Content and position information are intermingled throughout the network. While ideally the network would learn to balance the different types of information, in practice it does not. In this section we show that factoring the model to explicitly separate content and position information results in increased parsing accuracy. 

To help gauge the relative importance of the two types of attention, we trained a modified version of our model that was only allowed to use position attention. This constraint was enforced by making the query and key vectors used for the attention mechanism be linear transformations of the corresponding word's position embedding: $Q^{(n)} = PW_Q^{(n)}$ and $K^{(n)} = PW_K^{(n)}$. The per-head weight matrices now multiply a matrix $P$ containing the same position embeddings that are used at the input to the encoder, rather than the layer input $X$ (as in Section~\ref{subsec:self-attention}). However, value vectors $V^{(n)}=XW_V^{(n)}$ remain unchanged and continue to carry content-related information.

We expected our parser to still achieve reasonable performance when restricted to only use positional attention because the resulting architecture can be viewed as a generalization of a multi-layer convolutional neural network. The 8 attention heads at each layer of our model can mimic the behavior of a size-8 convolutional filter, but can also determine their attention targets dynamically and need not respect any translation-invariance properties. Disabling content-based attention throughout all 8 layers of the network results in a development-set accuracy decrease of only 0.27 F1. While we expected reasonable parsing performance in this setting, it seems strange that content-based attention benefits our model to such a small degree.

We next investigate the possibility that intermingling content and position information in a single vector can cause one type of attention to dominate over the other and compromise the network's ability to find the optimal balance of the two. To do this we propose a factored version of our model that explicitly separates content and position information.

A first step is to replace the component-wise addition $z_t = w_t + m_t + p_t$ (where $w_t$, $m_t$, and $p_t$ represent word, tag, and position embeddings, respectively) with a concatenation $z_t = [w_t+m_t ; p_t]$. We preserve the size of the vector $z_t$ by cutting the dimensionality of embeddings in half for the concatenative scheme. However, simply isolating the position-related components of the input vectors in this manner does not improve the performance of our network: the concatenative network achieves a development-set F1 of 92.60 (not much different from 92.67 F1 using the model in Section~\ref{sec:base-model}).

\newcommand{\content}[1]{{#1}^{(c)}}
\newcommand{\position}[1]{{#1}^{(p)}}
The issue with intermingling information is not the component-wise addition per se. In fact, concatenation and addition often perform similarly in high dimensions (especially when the resulting vector is immediately multiplied by a matrix that intermingles the two sources of information). On that note, we can examine how the mixed vectors are used later in the network, and in particular in the query-key dot products for the attention mechanism. If we have a query-key dot product $q \cdot k$ (see Section~\ref{subsec:self-attention}) where we imagine $q$ decomposing into content and positional information as $q = \content{q} + \position{q}$ (and likewise for $k$), we have $q \cdot k = (\content{q} + \position{q}) \cdot (\content{k} + \position{k})$. This formulation includes cross-terms such as $\content{q} \cdot \position{k}$; for example it is possible to learn a network where the word \emph{the} always attends to the \emph{5th position} in the sentence. Such cross-attention seems of limited use compared to the potential for overfitting that it introduces.

\begin{figure}
  \centering
    \includegraphics[width=0.49\textwidth]{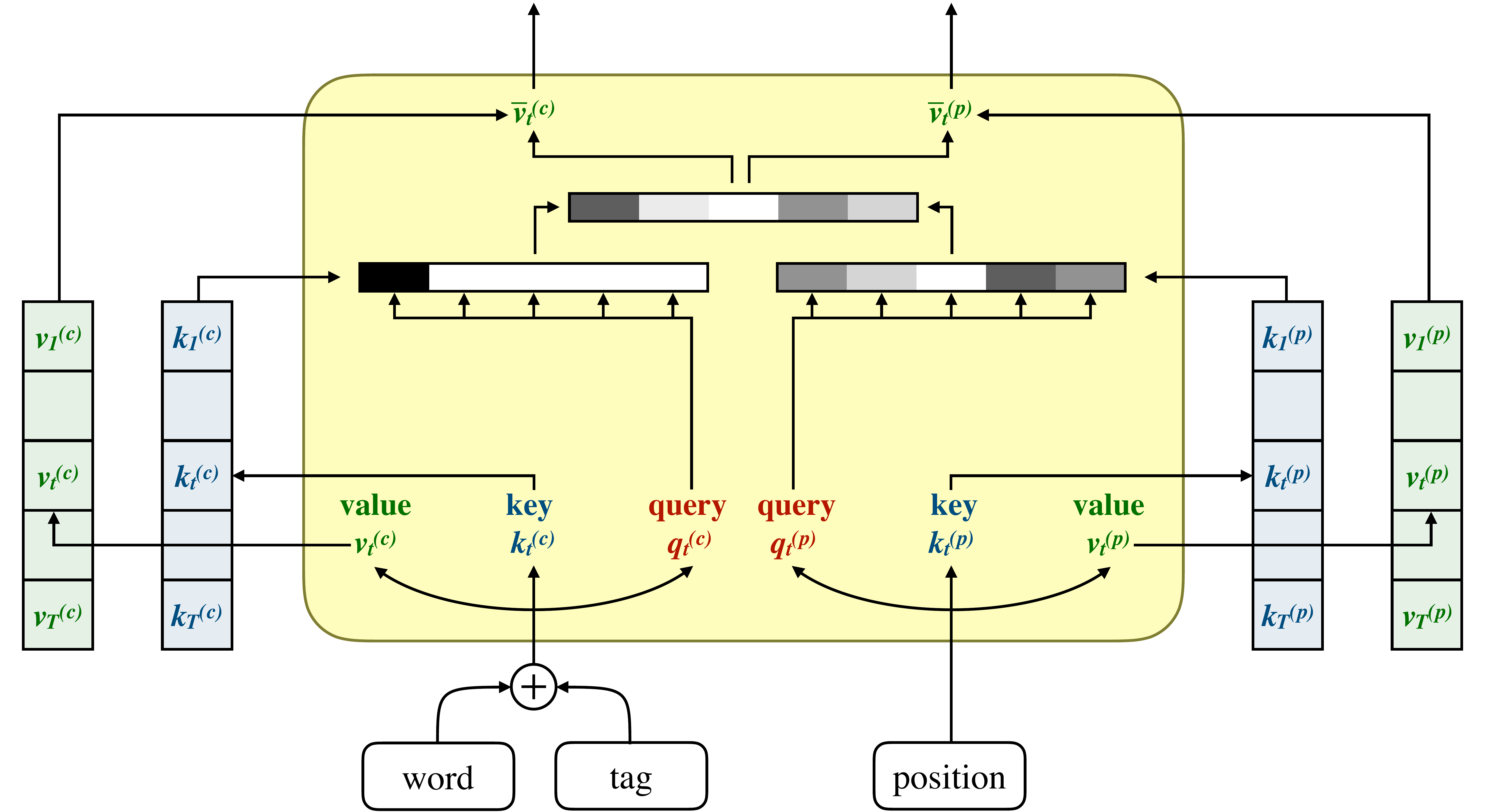}
    \caption{A single attention head, after factoring content and position information. Attention probabilities are calculated separately for the two types of information, and a combined probability distribution is then applied to both types of input information.}
    \label{fig:factored-head}
\end{figure}

To complete our factored model, we find all cases where a vector $x = [\content{x} ; \position{x}]$ is multiplied by a parameter matrix, and replace the matrix multiplication $c = Wx$ with a split form $c = [\content{c} ; \position{c}] = [\content{W}\content{x}; \position{W}\position{x}]$. This causes a number of intermediate quantities in our model to be factored, including all query and key vectors. Query-key dot products now decompose as $q \cdot k = \content{q} \cdot \content{k} + \position{q} \cdot \position{k}$. The result of factoring a single attention head, shown in Figure~\ref{fig:factored-head}, can also be viewed as separately applying attention to $\content{x}$ and $\position{x}$, except that the log-probabilities in the two halves are added together prior to value lookup.
The feed-forward sublayers in our model (Section~\ref{subsec:feed-forward}) are likewise split into two independent portions that operate on position and content information.

Alternatively, factoring can be seen as enforcing the block-sparsity constraint
$$W = \begin{bmatrix}\content{W} & 0 \\ 0 & \position{W}\end{bmatrix}$$
on parameter matrices throughout our model. We maintain the same vector sizes as in Section~\ref{sec:base-model}, which means that factoring strictly reduces the number of trainable parameters. For simplicity, we split each vector into equal halves that contain position and content information, cutting the number of model parameters roughly in half.
This factored scheme is able to achieve 93.15 development-set F1, an improvement of almost 0.5 F1 over the unfactored model.

These results suggest that factoring different types of information leads to a better parser, but there is in principle a confound: perhaps by making all matrices block-sparse we've stumbled across a better hyperparameter configuration. For example, these gains could be due to a difference in the number of trainable parameters alone. To control for this confound we also evaluated a version of our model that enforces block-sparsity throughout, but retains the use of component-wise addition at the inputs. This model achieves 92.63 F1 (not much different from the unfactored model), which supports our hypothesis that true factoring of information is important.

\section{Analysis of our Model}
\label{sec:analysis}

The defining feature of our encoder is the use of self-attention, which is the only mechanism for transfer of information between different locations throughout a sentence. The attention is further factored into types: content-based attention and position-based attention. In this section, we analyze the manner in which our model uses this attention mechanism to make its predictions.

\subsection{Content vs. Position Attention}

To examine the relative utilization of content-based vs. position-based attention in our architecture, we perturb a trained model at test-time by selectively zeroing out the contribution of either the content or the position component to any attention mechanism. This can be done independently at different layers; the results of this experiment are shown in Table~\ref{table:content-position-disable}.

\begin{table}[t!]
\begin{center}
\begin{tabular}{@{}llc@{}}
\toprule
\multicolumn{2}{c}{Attention}\\
\cmidrule{1-2}
Content & Position & F1\\
\midrule
All 8 layers & All 8 layers & 93.15 \\
All 8 layers & Disabled & 72.45 \\
Disabled & All 8 layers & 90.84 \\
First 4 layers only & All 8 layers & 91.77 \\
Last 4 layers only & All 8 layers & 92.82 \\
First 6 layers only & All 8 layers & 92.42 \\
Last 6 layers only & All 8 layers & 92.90 \\
\bottomrule
\end{tabular}
\end{center}
\caption{
\label{table:content-position-disable}
Development-set F1 scores when content and/or position attention is selectively disabled \emph{at test-time only} for a subset of the layers in our model. Position attention is the most important contributor to our model, but content attention is also helpful (especially at the final layers of the encoder).
}
\end{table}

We can see that our model learns to use a combination of the two attention types, with position-based attention being the most important. We also see that content-based attention is more useful at later layers in the network, which is consistent with the idea that the initial layers of our model act similarly to a dilated convolutional network while the upper layers have a greater balance between the two attention types.

\subsection{Windowed Attention}

\begin{table}[t!]
\begin{center}
\begin{tabular}{@{}lcc@{}}
\toprule
Distance & \phantom{a}F1 (strict)\phantom{a} & F1 (relaxed) \\
\midrule
5  & 81.65 & 89.82 \\
10 & 89.83 & 92.20 \\
15 & 91.72 & 92.78 \\
20 & 92.48 & 92.91 \\
30 & 93.01 & 93.09 \\
40 & 93.04 & 93.12 \\
$\infty$ & \multicolumn{2}{c}{93.15} \\
\bottomrule
\end{tabular}
\end{center}
\caption{
\label{table:distance-pretrained}
Development-set F1 scores when attention is constrained to not exceed a particular distance in the sentence \emph{at test time only}. In the \emph{relaxed} setting, the first and last two tokens of the sentence can attend to any word and be attended to by any word, to allow for sentence-wide pooling of information.
}
\end{table}

We can also examine our model's use of long-distance context information by applying windowing to the attention mechanism. We begin by taking our trained model and windowing the attention mechanism at test-time only. As shown in Table~\ref{table:distance-pretrained}, strict windowing yields poor results: even a window of size 40 causes a loss in parsing accuracy compared to the original model. When we began to investigate \emph{how} the model makes use of long-distance attention, we immediately found that there are particular attention heads at some layers in our model that almost always attend to the start token. This suggests that the start token is being used as the location for some sentence-wide pooling/processing, or perhaps as a dummy target location when a head fails to find the particular phenomenon that it's learned to search for. In light of this observation, we introduce a \emph{relaxed} variation on the windowing scheme, where the start token, first word, last word, and stop token can participate in all possible uses of attention, but pairs of other words in the sentence can only attend to each other if they are within a given window. We include three other positions in addition to the start token to do our best to cover possible locations for global pooling by our model. Results for relaxed windowing at test-time only are also shown in Table~\ref{table:distance-pretrained}. Even when we allow global processing to take place at designated locations such as the start token, our model is able to make use of long-distance dependencies at up to length 40.

\begin{table}[t!]
\begin{center}
\begin{tabular}{@{}lcc@{}}
\toprule
Distance & \phantom{a}F1 (strict)\phantom{a} & F1 (relaxed) \\
\midrule
5  & 92.74 & 92.94 \\
10 & 92.92 & 93.00 \\
20 & 93.06 & 93.17 \\
$\infty$ & \multicolumn{2}{c}{93.15} \\
\bottomrule
\end{tabular}
\end{center}
\caption{
\label{table:distance-newtrained}
Development-set F1 scores when attention is constrained to not exceed a particular distance in the sentence \emph{during training and at test time}. In the \emph{relaxed} setting, the first and last two tokens of the sentence can attend to any word and be attended to by any word, to allow for sentence-wide pooling of information.
}
\end{table}

Next, we examine whether the parser's use of long-distance dependencies is actually essential to performing the task by retraining our model subject to windowing. To evaluate the role of global computation, we consider both strict and relaxed windowing. In principle we could have replaced relaxed windowing at training time with explicit provisions for global computation, but for analysis purposes we choose to minimize departures from our original architecture.

The results, shown in Table~\ref{table:distance-newtrained}, demonstrate that long-distance dependencies continue to be essential for achieving maximum parsing accuracy using our model. Note that when a window of size 10 was imposed at training time, this was \emph{per-layer} and the series of 8 layers actually had an effective context size of around \emph{80} -- which was still insufficient to recover the performance of our full parser (with either approach to windowing). The side-by-side comparison of strict and relaxed windowing shows that the ability to pool global information, using the designated locations that are always available in the relaxed scheme, consistently translates to accuracy gains but is insufficient to compensate for small window sizes. This suggests that not only must the information signal from long-distance tokens be available in principle, but that it also helps to have this information be directly accessible without an intermediate bottleneck.

\section{Lexical Models}
\label{sec:lexical}

The models described in previous sections all rely on pretagged input sentences, where the tags are predicted using the Stanford tagger. We use the same pretagged dataset as \citet{cross_span-based_2016}.
In this section we explore two alternative classes of lexical models: those that use no external systems or data of any kind, as well as word vectors that are pretrained in an unsupervised manner.

\subsection{Models with Subword Features}
\label{subsec:morpho}

If tag embeddings are removed from our model and only word embeddings remain (where word embeddings are learned jointly with other model parameters), performance suffers by around 1 F1. To restore performance without introducing any dependencies on an external system, we explore incorporating lexical features directly into our model. The results for different approaches we describe in this section are shown in Table~\ref{table:replace-tags}.

\begin{table}[t!]
\begin{center}
\begin{tabular}{@{}lcc@{}}
\toprule
& \multicolumn{2}{c}{Word embeddings} \\
\cmidrule{2-3}
& \cmark & \xmark \\
\midrule
None & 92.20 & -- \\
Tags & 93.15 & -- \\
CharLSTM & 93.40 & 93.61 \\
CharConcat & 93.32 & 93.35 \\
\bottomrule
\end{tabular}
\end{center}
\caption{\label{table:replace-tags}
Development-set F1 scores for different approaches to handling morphology, with and without the addition of learned word embeddings.
}
\end{table}

We first evaluate an approach (\textsc{CharLSTM}) that independently runs a bidirectional LSTM over the characters in each word and uses the LSTM outputs in place of part-of-speech tag embeddings. We find that this approach performs better than using predicted part-of-speech tags. We can further remove the word embeddings (leaving the character LSTMs only), which does not seem to hurt and can actually help increase parsing accuracy.

Next we examine the importance of recurrent connections by constructing and evaluating a simpler alternative. Our approach (\textsc{CharConcat}) is inspired by \citet{hall_less_2014}, who found it effective to replace words with frequently-occurring suffixes, and the observation that our original tag embeddings are rather high-dimensional. To represent a word, we extract its first 8 letters and last 8 letters, embed each letter, and concatenate the results. If we use 32-dimensional embeddings, the 16 letters can be packed into a 512-dimensional vector -- the same size as the inputs to our model. This size for the inputs in our model was chosen to simplify the use of residual connections (by matching vector dimensions), even though the inputs themselves could have been encoded in a smaller vector. This allows us to directly replace tag embeddings with the 16-letter prefix/suffix concatenation. For short words, embeddings of a padding token are inserted as needed. Words longer than 16 letters are represented in a lossy manner by this concatenative approach, but we hypothesize that prefix/suffix information is enough for our task. We find this simple scheme remarkably effective: it is able to outperform pretagging and can operate even in the absence of word embeddings. However, its performance is ultimately not quite as good as using a character LSTM.

Given the effectiveness of the self-attentive encoder at the sentence level, it is aesthetically appealing to consider it as a sub-word architecture as well.  However, it was empirically much slower, did not parallelize better than a character-level LSTM (because words tend to be short), and initial results underperformed the LSTM. One explanation is that in a lexical model, one only wants to compute a single vector per word, whereas the self-attentive architecture is better adapted for producing context-aware summaries at multiple positions in a sequence.

\subsection{External Embeddings}
\label{subsec:elmo}

Next, we consider a version of our model that uses external embeddings. Recent work by \citet{peters_deep_2018} has achieved state-of-the-art performance across a range of NLP tasks by augmenting existing models with a new technique for word representation called ELMo (Embeddings from Language Models). Their approach is able to capture both subword information and contextual clues: the embeddings are produced by a network that takes characters as input and then uses an LSTM to capture contextual information when producing a vector representation for each word in a sentence.

We evaluate a version of our model that uses ELMo as the sole lexical representation, using publicly available ELMo weights. These pre-trained word representations are 1024-dimensional, whereas all of our factored models thus far have 512-dimensional content representations; we found that the most effective way to address this mismatch is to project the ELMo vectors to the required dimensionality using a learned weight matrix. With the addition of contextualized word representations, we hypothesized that a full 8 layers of self-attention would no longer be necessary. This proved true in practice: our best development set result of 95.21 F1 was obtained with a 4-layer encoder.

\section{Results}
\label{sec:results}

\subsection{English (WSJ)}

\begin{table}[t!]
\begin{center}
\begin{tabular}{@{}lcc@{}}
\toprule
Encoder Architecture & F1 (dev) & $\Delta$ \\
\midrule
LSTM \citep{gaddy_analysis_2018} & 92.24 & -0.43 \\
Self-attentive (Section~\ref{sec:base-model}) & 92.67 & \phantom{-}0.00 \\
+ Factored (Section~\ref{sec:factored-model}) & 93.15 & \phantom{-}0.48\\
+ CharLSTM (Section~\ref{subsec:morpho}) & 93.61 & \phantom{-}0.94\\
+ ELMo (Section~\ref{subsec:elmo}) & 95.21 & \phantom{-}2.54\\
\bottomrule
\end{tabular}
\end{center}
\caption{\label{table:wsj-recap} A comparison of different encoder architectures and their development-set performance relative to our base self-attentive model.}
\end{table}

\begin{table}[t!]
\begin{center}
\begin{tabular}{@{}lccc@{}}
\toprule
& LR & LP & F1  \\
\midrule
\textbf{Single model, WSJ only\hspace{-1em}} \\
\addlinespace
\citet{vinyals_grammar_2015} & -- & -- & 88.3\phantom{0} \\
\citet{cross_span-based_2016} & 90.5\phantom{0} & 92.1\phantom{0}  & 91.3\phantom{0} \\
\citet{gaddy_analysis_2018} & 91.76 & 92.41 & 92.08 \\
\citet{stern_effective_2017} & 92.57 & 92.56 & 92.56 \\
Ours (CharLSTM) & \textbf{93.20} &\textbf{93.90} & \textbf{93.55} \\
\addlinespace
\textbf{Multi-model/External} \\
\addlinespace
\citet{durrett_neural_2015} & -- & -- & 91.1\phantom{0} \\
\citet{vinyals_grammar_2015} & -- & -- & 92.8\phantom{0} \\
\citet{dyer_recurrent_2016} & -- & -- & 93.3\phantom{0} \\
\citet{choe_parsing_2016}\hspace{-1em} & -- & -- & 93.8\phantom{0} \\
\citet{liu_in_order_2017} & -- & -- & 94.2\phantom{0} \\
\citet{fried_improving_2017} & -- & -- & 94.66 \\
Ours (ELMo) & 94.85 & 95.40 & \textbf{95.13} \\
\bottomrule
\end{tabular}
\end{center}
\caption{\label{table:wsj-test} Comparison of F1 scores on the WSJ test set.}
\end{table}

\begin{table*}
\centering
\scriptsize
\begin{tabular}{@{}lcccccccccc@{}}
\toprule
  &Arabic &Basque &French &German &Hebrew &Hungarian &Korean &Polish &Swedish &Avg \\
  \midrule
  \textbf{Dev (all lengths)} \\ \addlinespace
  \citet{coavoux_multilingual_2017} &
83.07&88.35&82.35&88.75&90.34&91.22&86.78\rlap{$^b$}&\textbf{94.0\phantom{0}}&79.64&87.16\\
  Ours (CharLSTM only) & \textbf{85.94} & \textbf{90.05} &84.27&91.26&90.50&92.23&\textbf{87.90}&93.94&79.34& \textbf{88.38} \\
  Ours (CharLSTM + word embeddings) & 85.59 & 89.31 &\textbf{84.42}&\textbf{91.39}&\textbf{90.78}&\textbf{92.32}&87.62&93.76&\textbf{79.71}&   88.32 \\
  \midrule
  \textbf{Test (all lengths)} \\ \addlinespace 
  \citet{bjorkelund_ims-wroclaw-szeged-cis_2014}, ensemble &
81.32\rlap{$^a$}&88.24&82.53&81.66&89.80&91.72&83.81&90.50&\textbf{85.50}&86.12 \\
  \citet{cross_span-based_2016} &--&--&83.31&--&--&--&--&--&--&--\\
  \citet{coavoux_multilingual_2017} &
82.92\rlap{$^b$}&88.81&82.49&85.34&89.87&92.34&86.04&93.64&84.0\phantom{0}&87.27\\
  Ours (model selected on dev) &\textbf{85.61}&\textbf{89.71}&\textbf{84.06}&\textbf{87.69}&\textbf{90.35}&\textbf{92.69}&\textbf{86.59}&\textbf{93.69}&84.45&\textbf{88.32}\\ \addlinespace 
  $\Delta$: Ours - Best Previous &\textbf{+2.69}&\textbf{+0.90}&\textbf{+0.75}&\textbf{+2.35}&\textbf{+0.48}&\textbf{+0.35}&\textbf{+0.55}&\textbf{+0.05}&-1.05 \\
  \bottomrule
\end{tabular}
\caption{\label{table:spmrl}Results on the SPMRL dataset. All values are F1 scores calculated using the version of \texttt{evalb} distributed with the shared task. $^a$\citet{bjorkelund_re_2013} $^b$Uses character LSTM, whereas other results from \citet{coavoux_multilingual_2017} use predicted part-of-speech tags.}
\end{table*}

The development set scores of the parser variations presented in previous sections are summarized in Table~\ref{table:wsj-recap}. Our best-performing parser used a factored self-attentive encoder over ELMo word representations.

The results of evaluating our model on the test set are shown in Table~\ref{table:wsj-test}. The test score of 93.55 F1 for our CharLSTM parser exceeds the previous best numbers for single-system parsers trained on the Penn Treebank (without the use of any external data, such as pre-trained word embeddings). When our parser is augmented with ELMo word representations, it achieves a new state-of-the-art score of 95.13 F1 on the WSJ test set.

Our WSJ-only parser took 18 hours to train using a single Tesla K80 GPU and can parse the 1,700-sentence WSJ development set in 8 seconds. When using ELMo embeddings, training time was 13 hours (not including the time needed to pre-train the word embeddings) and parsing the development set takes 24 seconds. Training and inference times are dominated by neural network computations; our single-threaded Cython implementation of the chart decoder (Section~\ref{subsec:decoder}) consumes a negligible fraction of total running time.

\subsection{Multilingual (SPMRL)}

We tested our model's ability to generalize across languages by training it on the nine languages represented in the SPMRL 2013/2014 shared tasks \citep{seddah_overview_2013}. To verify that our lexical representations can function for morphologically-rich languages and smaller treebanks, we restricted ourselves to running a subset of the exact models that we evaluated on English. In particular, we evaluated the model that uses a character-level LSTM, with and without the addition of learned word embeddings.
We did not evaluate ELMo in the multilingual setting because pre-trained ELMo weights were only available for English. Hyperparameters were unchanged compared to the English model with the exception of the learning rate, which we adjusted for some of the smaller datasets in the SPMRL task (see Appendix~\ref{sec:hyperparams}). Results are shown in Table~\ref{table:spmrl}.

Development set results show that the addition of word embeddings to a model that uses a character LSTM has a mixed effect: it improves performance for some languages, but hurts for others. For each language, we selected the trained model that performed better on the development set and evaluated it on the test set. On 8 of the 9 languages, our test set result exceeds the previous best-published numbers from any system we are aware of. The exception is Swedish, where the model of \citet{bjorkelund_ims-wroclaw-szeged-cis_2014} continues to be state-of-the-art despite a number of approaches proposed in the intervening years that have achieved better performance on other languages. We note that their model uses ensembling (via product grammars) and a reranking step, whereas our model was only evaluated in the single-system condition.

\section{Conclusion}

In this paper, we show that the choice of encoder can have a substantial effect on parser performance.  In particular, we demonstrate state-of-the-art parsing results with a novel encoder based on factored self-attention. The gains we see come not only from incorporating \emph{more} information (such as subword features or externally-trained word representations), but also from structuring the architecture to \emph{separate} different kinds of information from each other. Our results suggest that further research into different ways of encoding utterances can lead to additional improvements in both parsing and other natural language processing tasks.

\section*{Acknowledgments}

NK is supported by an NSF Graduate Research Fellowship. This research used the Savio computational cluster provided by the Berkeley Research Computing program at the University of California, Berkeley.


\bibliography{acl2018}

\begin{thebibliography}{18}
\expandafter\ifx\csname natexlab\endcsname\relax\def\natexlab#1{#1}\fi

\bibitem[{Ba et~al.(2016)Ba, Kiros, and Hinton}]{ba_layer_2016}
Jimmy~Lei Ba, Jamie~Ryan Kiros, and Geoffrey~E. Hinton. 2016.
\newblock \href {http://arxiv.org/abs/1607.06450} {Layer {Normalization}}.
\newblock \emph{arXiv:1607.06450 [cs, stat]}.
\newblock ArXiv: 1607.06450.

\bibitem[{Bj\"{o}rkelund et~al.(2014)Bj\"{o}rkelund, Cetinoglu, Fale\'{n}ska,
  Farkas, Mueller, Seeker, and
  Sz\'{a}nt\'{o}}]{bjorkelund_ims-wroclaw-szeged-cis_2014}
Anders Bj\"{o}rkelund, Ozlem Cetinoglu, Agnieszka Fale\'{n}ska, Rich\'{a}rd
  Farkas, Thomas Mueller, Wolfgang Seeker, and Zsolt Sz\'{a}nt\'{o}. 2014.
\newblock The {IMS}-{Wrocław}-{Szeged}-{CIS} entry at the {SPMRL} 2014 shared
  task: {Reranking} and morphosyntax meet unlabeled data.
\newblock In \emph{Proceedings of the {First} {Joint} {Workshop} on
  {Statistical} {Parsing} of {Morphologically} {Rich} {Languages} and
  {Syntactic} {Analysis} of {Non}-{Canonical} {Languages}}, pages 97--102.

\bibitem[{Bj\"{o}rkelund et~al.(2013)Bj\"{o}rkelund, Cetinoglu, Farkas,
  Mueller, and Seeker}]{bjorkelund_re_2013}
Anders Bj\"{o}rkelund, Ozlem Cetinoglu, Rich\'{a}rd Farkas, Thomas Mueller, and
  Wolfgang Seeker. 2013.
\newblock \href {http://www.aclweb.org/anthology/W13-4916} {({Re})ranking meets
  morphosyntax: State-of-the-art results from the {SPMRL} 2013 shared task}.
\newblock In \emph{Proceedings of the Fourth Workshop on Statistical Parsing of
  Morphologically-Rich Languages}, pages 135--145, Seattle, Washington, USA.
  Association for Computational Linguistics.

\bibitem[{Choe and Charniak(2016)}]{choe_parsing_2016}
Do~Kook Choe and Eugene Charniak. 2016.
\newblock \href {https://doi.org/10.18653/v1/D16-1257} {Parsing as language
  modeling}.
\newblock In \emph{Proceedings of the 2016 Conference on Empirical Methods in
  Natural Language Processing}, pages 2331--2336. Association for Computational
  Linguistics.

\bibitem[{Coavoux and Crabb{\'e}(2017)}]{coavoux_multilingual_2017}
Maximin Coavoux and Benoit Crabb{\'e}. 2017.
\newblock \href {http://aclweb.org/anthology/E17-2053} {Multilingual
  lexicalized constituency parsing with word-level auxiliary tasks}.
\newblock In \emph{Proceedings of the 15th Conference of the European Chapter
  of the Association for Computational Linguistics: Volume 2, Short Papers},
  pages 331--336. Association for Computational Linguistics.

\bibitem[{Cross and Huang(2016)}]{cross_span-based_2016}
James Cross and Liang Huang. 2016.
\newblock \href {https://doi.org/10.18653/v1/D16-1001} {Span-based constituency
  parsing with a structure-label system and provably optimal dynamic oracles}.
\newblock In \emph{Proceedings of the 2016 Conference on Empirical Methods in
  Natural Language Processing}, pages 1--11. Association for Computational
  Linguistics.

\bibitem[{Durrett and Klein(2015)}]{durrett_neural_2015}
Greg Durrett and Dan Klein. 2015.
\newblock \href {https://doi.org/10.3115/v1/P15-1030} {Neural {CRF} parsing}.
\newblock In \emph{Proceedings of the 53rd Annual Meeting of the Association
  for Computational Linguistics and the 7th International Joint Conference on
  Natural Language Processing (Volume 1: Long Papers)}, pages 302--312.
  Association for Computational Linguistics.

\bibitem[{Dyer et~al.(2016)Dyer, Kuncoro, Ballesteros, and
  Smith}]{dyer_recurrent_2016}
Chris Dyer, Adhiguna Kuncoro, Miguel Ballesteros, and Noah~A. Smith. 2016.
\newblock \href {https://doi.org/10.18653/v1/N16-1024} {Recurrent neural
  network grammars}.
\newblock In \emph{Proceedings of the 2016 Conference of the North American
  Chapter of the Association for Computational Linguistics: Human Language
  Technologies}, pages 199--209. Association for Computational Linguistics.

\bibitem[{Fried et~al.(2017)Fried, Stern, and Klein}]{fried_improving_2017}
Daniel Fried, Mitchell Stern, and Dan Klein. 2017.
\newblock \href {https://doi.org/10.18653/v1/P17-2025} {Improving neural
  parsing by disentangling model combination and reranking effects}.
\newblock In \emph{Proceedings of the 55th Annual Meeting of the Association
  for Computational Linguistics (Volume 2: Short Papers)}, pages 161--166.
  Association for Computational Linguistics.

\bibitem[{Gaddy et~al.(2018)Gaddy, Stern, and Klein}]{gaddy_analysis_2018}
David Gaddy, Mitchell Stern, and Dan Klein. 2018.
\newblock \href {https://arxiv.org/abs/1804.07853} {What's going on in neural
  constituency parsers? {An} analysis}.
\newblock In \emph{Proceedings of the 2018 Conference of the North American
  Chapter of the Association for Computational Linguistics: Human Language
  Technologies}. Association for Computational Linguistics.

\bibitem[{Hall et~al.(2014)Hall, Durrett, and Klein}]{hall_less_2014}
David Hall, Greg Durrett, and Dan Klein. 2014.
\newblock \href {http://www.anthology.aclweb.org/P/P14/P14-1022.pdf} {Less
  grammar, more features}.
\newblock In \emph{Proceedings of the 52nd {Annual} {Meeting} of the
  {Association} for {Computational} {Linguistics}}, volume~1, pages 228--237.

\bibitem[{Liu and Zhang(2017)}]{liu_in_order_2017}
Jiangming Liu and Yue Zhang. 2017.
\newblock \href {http://aclweb.org/anthology/Q17-1029} {In-order
  transition-based constituent parsing}.
\newblock \emph{Transactions of the Association for Computational Linguistics},
  5:413--424.

\bibitem[{Peters et~al.(2018)Peters, Neumann, Iyyer, Gardner, Clark, Lee, and
  Zettlemoyer}]{peters_deep_2018}
Matthew~E. Peters, Mark Neumann, Mohit Iyyer, Matt Gardner, Christopher Clark,
  Kenton Lee, and Luke Zettlemoyer. 2018.
\newblock \href {http://arxiv.org/abs/1802.05365} {Deep contextualized word
  representations}.
\newblock In \emph{Proceedings of the 2018 Conference of the North American
  Chapter of the Association for Computational Linguistics: Human Language
  Technologies}. Association for Computational Linguistics.

\bibitem[{Seddah et~al.(2013)Seddah, Tsarfaty, K{\"u}bler, Candito, Choi,
  Farkas, Foster, Goenaga, Gojenola~Galletebeitia, Goldberg, Green, Habash,
  Kuhlmann, Maier, Nivre, Przepi{\'o}rkowski, Roth, Seeker, Versley, Vincze,
  Woli{\'{n}}ski, Wr{\'o}blewska, and de~la Clergerie}]{seddah_overview_2013}
Djam{\'e} Seddah, Reut Tsarfaty, Sandra K{\"u}bler, Marie Candito, Jinho~D.
  Choi, Rich{\'a}rd Farkas, Jennifer Foster, Iakes Goenaga, Koldo
  Gojenola~Galletebeitia, Yoav Goldberg, Spence Green, Nizar Habash, Marco
  Kuhlmann, Wolfgang Maier, Joakim Nivre, Adam Przepi{\'o}rkowski, Ryan Roth,
  Wolfgang Seeker, Yannick Versley, Veronika Vincze, Marcin Woli{\'{n}}ski,
  Alina Wr{\'o}blewska, and Eric~Villemonte de~la Clergerie. 2013.
\newblock \href {http://www.aclweb.org/anthology/W13-4917} {Overview of the
  {SPMRL} 2013 shared task: A cross-framework evaluation of parsing
  morphologically rich languages}.
\newblock In \emph{Proceedings of the Fourth Workshop on Statistical Parsing of
  Morphologically-Rich Languages}, pages 146--182. Association for
  Computational Linguistics.

\bibitem[{Stern et~al.(2017{\natexlab{a}})Stern, Andreas, and
  Klein}]{stern_minimal_2017}
Mitchell Stern, Jacob Andreas, and Dan Klein. 2017{\natexlab{a}}.
\newblock \href {https://doi.org/10.18653/v1/P17-1076} {A minimal span-based
  neural constituency parser}.
\newblock In \emph{Proceedings of the 55th Annual Meeting of the Association
  for Computational Linguistics (Volume 1: Long Papers)}, pages 818--827.
  Association for Computational Linguistics.

\bibitem[{Stern et~al.(2017{\natexlab{b}})Stern, Fried, and
  Klein}]{stern_effective_2017}
Mitchell Stern, Daniel Fried, and Dan Klein. 2017{\natexlab{b}}.
\newblock \href {http://aclweb.org/anthology/D17-1178} {Effective inference for
  generative neural parsing}.
\newblock In \emph{Proceedings of the 2017 Conference on Empirical Methods in
  Natural Language Processing}, pages 1695--1700. Association for Computational
  Linguistics.

\bibitem[{Vaswani et~al.(2017)Vaswani, Shazeer, Parmar, Uszkoreit, Jones,
  Gomez, Kaiser, and Polosukhin}]{vaswani_attention_2017}
Ashish Vaswani, Noam Shazeer, Niki Parmar, Jakob Uszkoreit, Llion Jones,
  Aidan~N Gomez, {\L}ukasz Kaiser, and Illia Polosukhin. 2017.
\newblock \href
  {http://papers.nips.cc/paper/7181-attention-is-all-you-need.pdf} {Attention
  is all you need}.
\newblock In I.~Guyon, U.~V. Luxburg, S.~Bengio, H.~Wallach, R.~Fergus,
  S.~Vishwanathan, and R.~Garnett, editors, \emph{Advances in Neural
  Information Processing Systems 30}, pages 5998--6008. Curran Associates, Inc.

\bibitem[{Vinyals et~al.(2015)Vinyals, Kaiser, Koo, Petrov, Sutskever, and
  Hinton}]{vinyals_grammar_2015}
Oriol Vinyals, Łukasz Kaiser, Terry Koo, Slav Petrov, Ilya Sutskever, and
  Geoffrey Hinton. 2015.
\newblock \href
  {http://papers.nips.cc/paper/5635-grammar-as-a-foreign-language.pdf} {Grammar
  as a foreign language}.
\newblock In \emph{Advances in {Neural} {Information} {Processing} {Systems}
  28}, pages 2755--2763. Curran Associates, Inc.

\end{thebibliography}
\bibliographystyle{acl_natbib}

\appendix
\newpage
\phantom{No text here}

\begin{table*}[ht]
\begin{center}
\begin{tabular}{@{}lp{0.8\textwidth}l@{}}
\toprule
Symbol & Description & Value \\
\midrule
$N$ & Number of layers (when not using ELMo embeddings) & 8 \\
$N$ & Number of layers (when using ELMo embeddings) & 4 \\
$d_{model}$ & Model dimensionality & 1024 \\
$h$ & Number of attention heads & 8 \\
$d_k$ & Size of attention query/key vectors & 64 \\
$d_v$ & Size of attention value vectors & 64 \\
$d_{ff}$ & Size of intermediate vectors in the feed-forward sublayer & 2048 \\
& Size of character embeddings (CharConcat) & 32 \\
& Size of character embeddings (CharLSTM) & 64 \\
\midrule
& Attention dropout probability; see \citet{vaswani_attention_2017} & 0.2 \\
& ReLU dropout probability in feed-forward sublayer & 0.1 \\
& Residual dropout probability (at all residual connections) & 0.2 \\
& Word embedding dropout probability & 0.4 \\
& Dropout probability for part-of-speech tag embeddings& 0.2 \\
& Dropout probability for CharConcat/CharLSTM morphological representations & 0.2 \\
& Character embedding dropout probability at the inputs to CharLSTM & 0.2 \\
\bottomrule
\end{tabular}
\end{center}
\caption{
\label{table:hyperparams}
Model hyperparameters used for all of our experiments.
}
\end{table*}
\newpage

\section{Training Details}
\label{sec:hyperparams}

\subsection{Model Hyperparameters}
The hyperparameters for our model are shown in Table~\ref{table:hyperparams}. Hyperparameters were tuned on the development set for English.

\subsection{Optimizer Parameters}

Our model was trained using Adam with a batch size of 250 sentences. For the first 160 batches (equal to 1 epoch for English), the learning rate was increased linearly from $0$ up to the base learning rate shown in Table~\ref{table:lr}. Development-set performance was evaluated four times per epoch; if it did not improve for 5 epochs in a row the learning rate was halved. The iterate that performed best on the development set was taken as the output of the training procedure.

To ensure stability of the optimizer, we found it important to use a large batch size, to warm up the learning rate over time (similar to \citet{vaswani_attention_2017}), and to pick an appropriate learning rate.

\subsection{Position Embeddings}

All variations of our model use learned position embeddings. Our attempts to use the sinusoidal position embeddings proposed by \citet{vaswani_attention_2017} consistently performed worse than using learned embeddings.

\begin{table}
\begin{center}
\begin{tabular}{@{}ll@{}}
\toprule
Language & Base Learning Rate \\
\midrule
English & $0.0008$ \\
Hebrew & $0.002$ \\
Polish & $0.0015$ \\
Swedish & $0.002$ \\
All others & $0.0008$ \\
\bottomrule
\end{tabular}
\end{center}
\caption{
\label{table:lr}
Learning rates.
}
\end{table}

\end{document}